\documentclass[fleqn,10pt]{wlpeerj}

\usepackage{booktabs}
\usepackage{multirow}
\usepackage{amsmath}
\usepackage{graphicx}
\usepackage[section]{placeins}
\graphicspath{{figures/}}

\title{FireSenseNet: A Dual-Branch CNN with Cross-Attentive Feature Interaction for Next-Day Wildfire Spread Prediction}

\author[1]{Jinzhen Han}
\author[1]{JinByeong Lee}
\author[2]{Hak Han}
\author[3]{YeonJu Na}
\author[3]{Jae-Joon Lee}
\affil[1]{Department of Civil, Architectural \& Environment Engineering, Sungkyunkwan University, Suwon, South Korea}
\affil[2]{AI Urban Disaster Prevention Research Center, Advanced Institute of Convergence Technology (AICT), Suwon 16229, Republic of Korea}
\affil[3]{Department of Fire Safety Engineering, Jeonju University, Republic of Korea}
\corrauthor[3]{Jae-Joon Lee}{safety\_lee@jj.ac.kr}

\begin{abstract}
Accurate prediction of next-day wildfire spread is critical for disaster response and resource allocation. Existing deep learning approaches typically concatenate heterogeneous geospatial inputs into a single tensor, ignoring the fundamental physical distinction between static fuel/terrain properties and dynamic meteorological conditions. We propose FireSenseNet, a dual-branch convolutional neural network equipped with a novel Cross-Attentive Feature Interaction Module (CAFIM) that explicitly models the spatially varying interaction between fuel and weather modalities through learnable attention gates at multiple encoder scales. Through a systematic comparison of seven architectures---spanning pure CNNs, Vision Transformers, and hybrid designs---on the Google Next-Day Wildfire Spread benchmark, we demonstrate that FireSenseNet achieves an F1 of 0.4176 and AUC-PR of 0.3435, outperforming all alternatives including a SegFormer with 3.8$\times$ more parameters (F1 = 0.3502). Ablation studies confirm that CAFIM provides a 7.1\% relative F1 gain over naive concatenation, and channel-wise feature importance analysis reveals that the previous-day fire mask dominates prediction while wind speed acts as noise at the dataset's coarse temporal resolution. We further incorporate Monte Carlo Dropout for pixel-level uncertainty quantification and present a critical analysis showing that common evaluation shortcuts inflate reported F1 scores by over 44\%.
\end{abstract}

\begin{document}

\flushbottom
\maketitle
\thispagestyle{empty}

%% ============================================================
\section*{Introduction}
%% ============================================================

Wildfires are among the most destructive natural hazards worldwide. In 2024 alone, the United States experienced over 64,000 wildfires burning more than 8.9 million acres \citep{NICC2024}. \citet{Iglesias2022} showed that U.S. fires became larger, more frequent, and more widespread throughout the 2000s, a trend expected to continue as climate change intensifies fire regimes. The demand for accurate, automated wildfire spread prediction has never been more urgent.

Deep learning has emerged as a promising paradigm for this task, enabling models to learn complex spatiotemporal patterns directly from multi-modal remote sensing observations \citep{DiGiuseppe2025, Huot2022}. However, a key limitation persists: existing architectures overwhelmingly treat heterogeneous geospatial inputs---static terrain, vegetation, and dynamic weather---as a flat concatenated tensor, ignoring the distinct physical processes that govern each modality. Fuel availability and topography constrain \emph{where} fire can spread; meteorological conditions determine \emph{how fast} and \emph{in which direction} it propagates. A model that encodes this distinction at the architectural level should outperform one that must discover it from data alone.

\begin{figure}[!htbp]
\centering
\includegraphics[width=\linewidth]{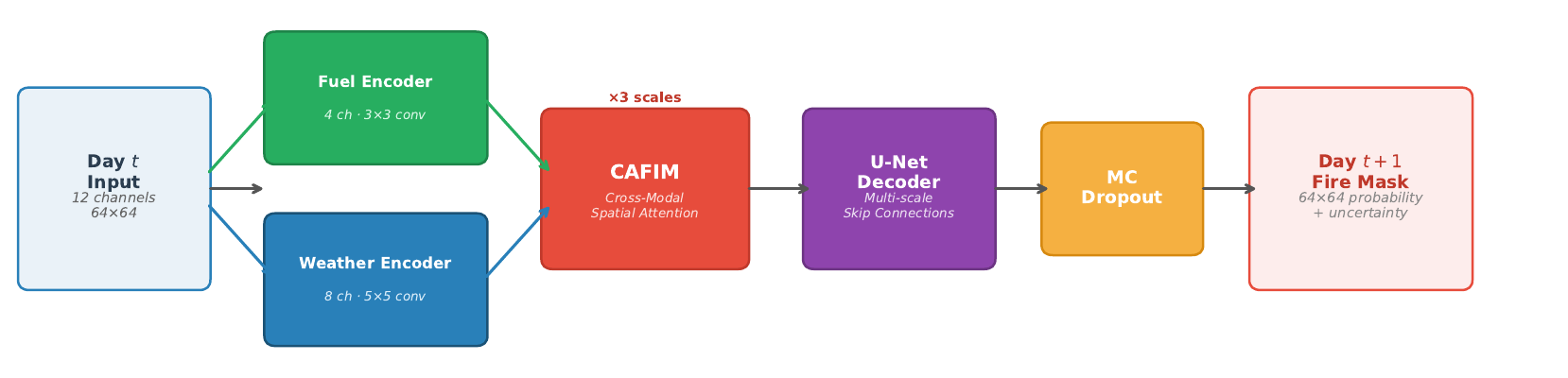}
\caption{FireSenseNet pipeline overview. Given 12-channel geospatial inputs from Day~$t$, dual-branch encoders with modality-specific kernel sizes process fuel/terrain (4\,ch, $3{\times}3$) and meteorological (8\,ch, $5{\times}5$) features independently. Cross-Attentive Feature Interaction Modules (CAFIM) fuse the two branches via learned spatial attention at three encoder scales. A U-Net decoder with multi-scale skip connections produces the Day~$t{+}1$ fire spread prediction, with Monte Carlo Dropout enabling pixel-level uncertainty estimation.}
\label{fig:overview}
\end{figure}

This paper proposes \textbf{FireSenseNet}, a dual-branch CNN architecture in which fuel/terrain features and meteorological features flow through independent encoder pathways before being fused via a novel \textbf{Cross-Attentive Feature Interaction Module (CAFIM)}---a learned spatial attention gate that determines, at each spatial location and encoder scale, how fuel and weather information should be combined (Figure~\ref{fig:overview}). A natural follow-up question is whether this domain-informed CNN design remains competitive against the Vision Transformers that have recently achieved strong results on semantic segmentation benchmarks \citep{Dosovitskiy2021, Xie2021}. Wildfire prediction, however, operates in a regime quite different from natural image segmentation: training sets are small ($\sim$15K samples), informative fire pixels are extremely sparse ($<$1\%), and the critical predictive signals are local rather than global. To answer this question rigorously, we conduct a systematic comparison across \textbf{seven architectures} spanning pure CNNs, pure Transformers, and hybrid designs, all trained under identical data pipelines and evaluation protocols.

Prior work on wildfire spread modeling spans a broad spectrum. Physics-based simulators such as FARSITE \citep{Finney1998} encode fire behavior equations governing heat transfer, fuel consumption, and wind-driven propagation, but require extensive parameter calibration and struggle with complex, nonlinear interactions at landscape scale \citep{Singh2025, Duff2021}. Statistical approaches learn empirical relationships from historical data yet typically assume linear or weakly nonlinear interactions and cannot capture the spatial structure of fire spread. Deep learning has advanced rapidly in this domain: early work applied neural networks to model fire spread dynamics \citep{Hodges2019} and assess fire susceptibility from remote sensing data \citep{Kantarcioglu2023}, while \citet{Huot2022} released the Next-Day Wildfire Spread benchmark and established a CNN autoencoder baseline (F1 = 0.359). Subsequent efforts explored deeper architectures \citep{Shadrin2024}, multimodal fusion strategies, and multi-temporal prediction \citep{Gerard2023}. \citet{Luo2026} proposed a U-Net variant with frequency-domain transforms reporting substantially higher F1 scores; however, as we demonstrate in our evaluation protocol analysis, such gains can be largely attributed to evaluation methodology rather than architectural innovation. Meanwhile, Vision Transformers \citep{Dosovitskiy2021} and efficient variants such as SegFormer \citep{Xie2021} and Swin Transformer \citep{Liu2021} have achieved strong results across semantic segmentation benchmarks and are increasingly adopted in remote sensing, yet they lack the inductive biases inherent to CNNs---translation equivariance and locality---which serve as strong priors when training data is limited \citep{Dosovitskiy2021}. A common limitation across all these approaches is the treatment of all input channels as a homogeneous feature tensor: none explicitly model the physical distinction between fuel/terrain properties and meteorological conditions---a distinction fundamental to fire behavior science.

Our contributions are as follows:

\begin{enumerate}[noitemsep]
    \item We propose \textbf{FireSenseNet}, a dual-branch CNN with Cross-Attentive Feature Interaction Modules (CAFIM) that achieves a new best F1 of 0.4176 on the Google Next-Day Wildfire Spread benchmark---surpassing the published CNN baseline of \citet{Huot2022} (F1 = 0.359) by 16.3\% and a SegFormer with 3.8$\times$ more parameters (F1 = 0.3502) by 19.2\%. Ablation confirms that CAFIM alone contributes a 7.1\% relative F1 gain over naive concatenation.
    \item We conduct the first \textbf{systematic architecture comparison} for this benchmark, evaluating seven designs across the CNN--Transformer spectrum under identical training conditions. Complementary analyses---channel-wise feature importance ablation, Monte Carlo Dropout uncertainty quantification, and a critical audit of evaluation protocols that inflate reported F1 by over 44\%---provide both practical insights and methodological guidance for the field.
\end{enumerate}

%% ============================================================
\section*{Dataset and Evaluation Protocol}
%% ============================================================

\subsection*{Dataset description}

We use the Next-Day Wildfire Spread dataset released by Google Research \citep{Huot2022}, comprising multi-modal geospatial rasters derived from satellite imagery, topographic data, and reanalysis weather products. Each sample is a $64 \times 64$ pixel patch with 12 input channels and a binary segmentation target representing the next-day fire extent.

The 12 input channels are organized into two semantic groups reflecting their physical roles in fire behavior:
\begin{itemize}[noitemsep]
    \item \textbf{Fuel/Terrain channels} (4): elevation, Normalized Difference Vegetation Index (NDVI), population density, and previous-day fire mask (PrevFireMask).
    \item \textbf{Meteorological channels} (8): wind direction (th), wind speed (vs), minimum temperature (tmmn), maximum temperature (tmmx), specific humidity (sph), precipitation (pr), Palmer Drought Severity Index (pdsi), and Energy Release Component (erc).
\end{itemize}

\begin{figure*}[!htbp]
\centering
\includegraphics[width=\textwidth]{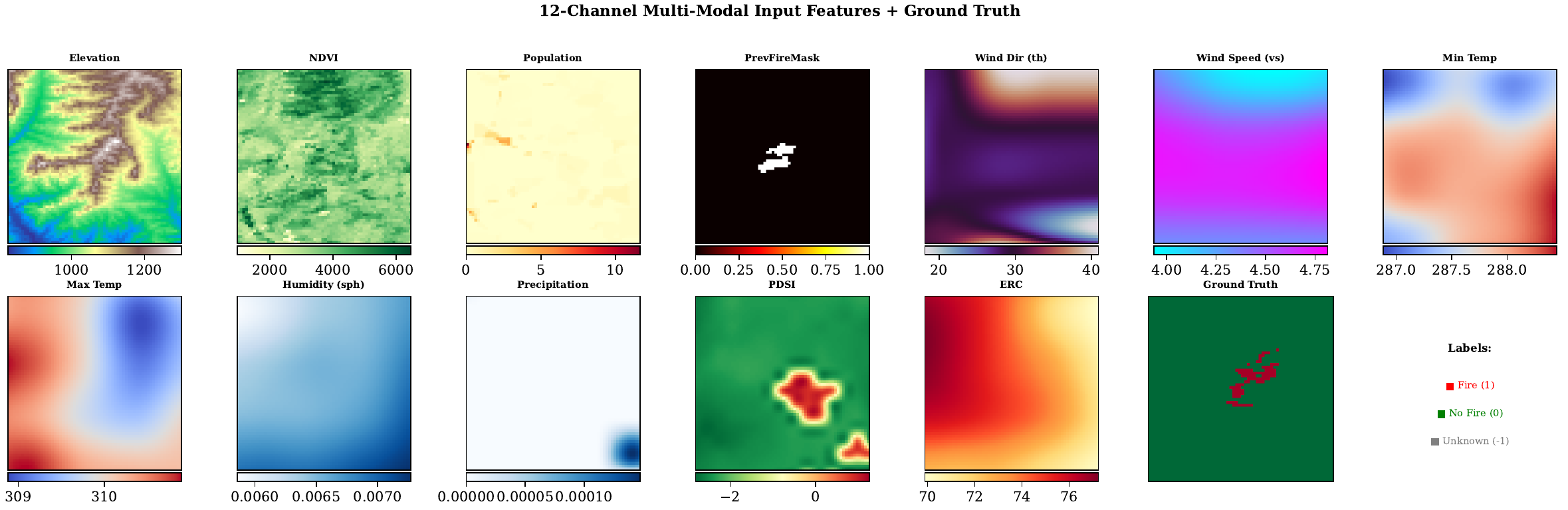}
\caption{Visualization of a representative test sample showing all 12 input channels and the ground truth fire mask. Static fuel/terrain features (top-left four panels) exhibit fine-grained spatial structure, while dynamic meteorological variables (remaining panels) vary smoothly across the patch. The ground truth (bottom-right) shows fire pixels (red), non-fire pixels (green), and unknown pixels (gray, labeled $-1$). This heterogeneity motivates the dual-branch design of FireSenseNet.}
\label{fig:12channel}
\end{figure*}

The dataset comprises approximately 15,000 training samples with validation and test splits in an 8:1:1 ratio. Figure~\ref{fig:12channel} visualizes a representative test sample, illustrating the heterogeneous spatial characteristics of these modalities: topographic features exhibit fine-grained structure, vegetation indices show patchy patterns, while meteorological variables vary smoothly---a diversity that motivates modality-specific processing.

A significant proportion of target pixels carry a label of $-1$, indicating regions where fire status is uncertain or unobservable. The treatment of these pixels has critical implications for evaluation integrity, as we discuss next.

\subsection*{Evaluation protocol and the problem of score inflation}

We adopt a strict evaluation protocol: the model receives Day $t$ features and predicts only the \emph{newly burned} area on Day $t+1$. All pixels with unknown labels ($-1$) are excluded from loss computation and metric evaluation. We report Precision, Recall, F1 score, and Area Under the Precision-Recall Curve (AUC-PR). An optimal classification threshold is selected by sweeping from 0.05 to 0.95 on the test set to maximize F1.

This protocol stands in contrast to practices observed in some prior work, where (i) the prediction target includes both the previous-day and next-day fire masks (``Both Days'' prediction), allowing the model to inflate its score by simply reproducing the PrevFireMask input; and (ii) unknown pixels ($-1$) are relabeled as non-fire (0), artificially increasing the true negative count. We quantify the impact of these shortcuts in Section: Results, demonstrating that they inflate F1 scores by 44--50\%.

All input channels are globally normalized using training-set statistics. Data augmentation consists of random horizontal and vertical flips during training.

%% ============================================================
\section*{Methods}
%% ============================================================

We investigate seven neural network architectures spanning a spectrum from pure CNN to pure Transformer, with hybrid designs in between. All architectures share the same data pipeline, loss function, and evaluation protocol to ensure fair comparison.

\subsection*{FireSenseNet: Dual-branch CNN with CAFIM}

FireSenseNet is grounded in the observation that fuel/terrain properties and meteorological conditions play distinct physical roles in fire spread: the former constrain the \emph{spatial envelope} of potential fire activity, while the latter modulate the \emph{rate and direction} of propagation within that envelope. Rather than forcing a single encoder to disentangle these roles, FireSenseNet processes them through independent pathways before fusing them via learned cross-modal attention.

\paragraph{Dual-branch encoder.} The fuel branch uses standard $3 \times 3$ residual blocks to capture the fine-grained spatial textures of terrain contours and vegetation boundaries. The weather branch employs larger $5 \times 5$ convolutional kernels, reflecting the broader spatial coherence of atmospheric fields. Both branches produce feature maps at three progressively downsampled scales ($64^2 \to 32^2 \to 16^2$).

\begin{figure}[!htbp]
\centering
\includegraphics[width=\linewidth]{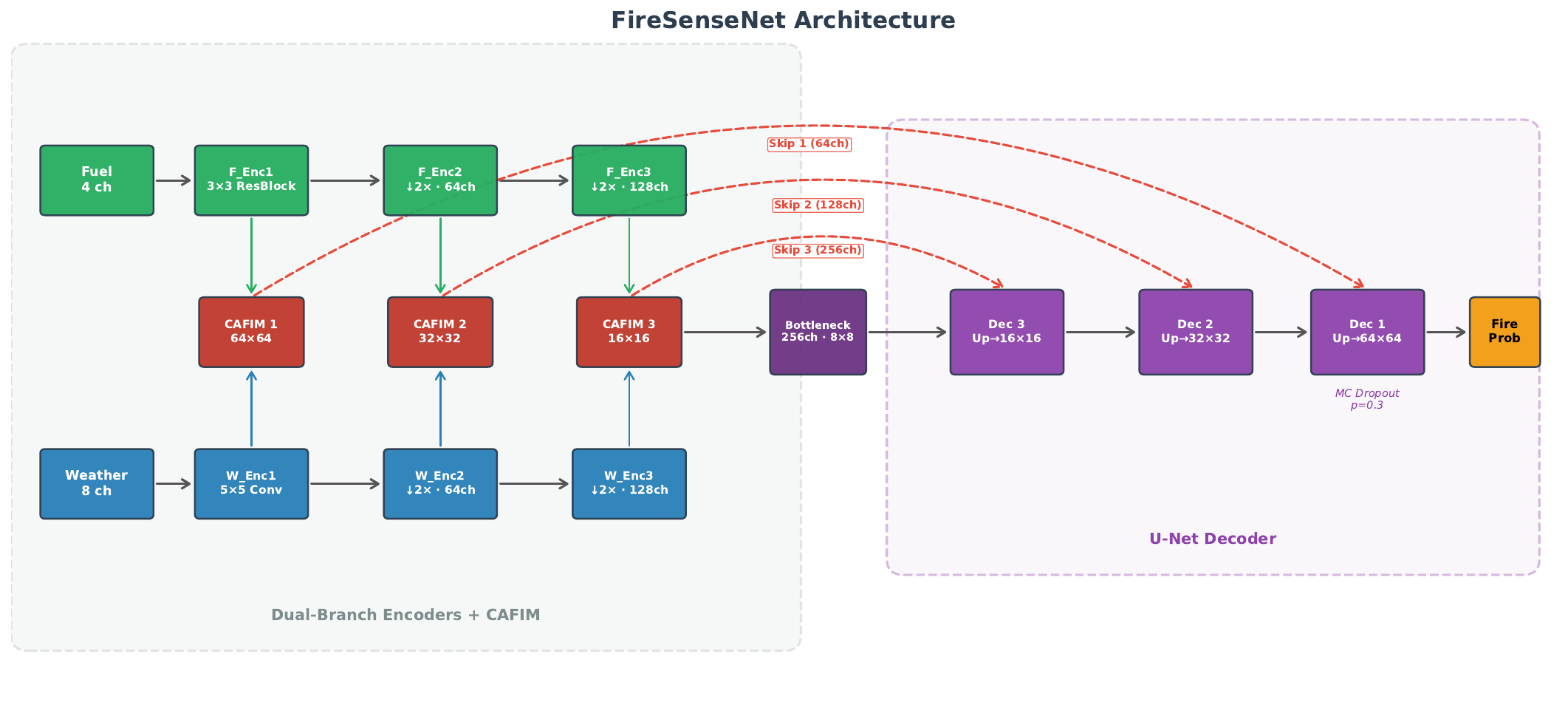}
\caption{FireSenseNet architecture. Fuel/terrain and meteorological inputs are processed by independent encoder branches with modality-appropriate kernel sizes. At each spatial scale, a Cross-Attentive Feature Interaction Module (CAFIM) generates a spatial attention map that gates the contribution of each modality before routing to the U-Net decoder. MC Dropout before the prediction head enables uncertainty quantification.}
\label{fig:architecture}
\end{figure}

\paragraph{Cross-Attentive Feature Interaction Module (CAFIM).} At each encoder scale, a CAFIM module fuses the two branches through a learned spatial attention gate (Figure~\ref{fig:architecture}). Given fuel features $F_\text{fuel}$ and weather features $F_\text{weather}$ at a given scale, CAFIM first projects each through a $1 \times 1$ convolution, concatenates the results, and generates a spatial attention map $\alpha \in [0,1]^{H \times W}$:

\begin{equation}
\alpha = \sigma\!\left(\text{Conv}_{3\times3}\!\left(\text{ReLU}\!\left(\text{Conv}_{3\times3}\!\left([W_f \cdot F_{\text{fuel}};\ W_w \cdot F_{\text{weather}}]\right)\right)\right)\right)
\end{equation}

where $W_f$ and $W_w$ are $1 \times 1$ projection convolutions and $\sigma$ denotes the sigmoid function. The fused output is a complementary gating:

\begin{equation}
F_{\text{fused}} = [\alpha \odot F_{\text{fuel}};\ (1-\alpha) \odot F_{\text{weather}}]
\end{equation}

This formulation encodes a physically meaningful prior: at each spatial location, the module learns whether the local fire dynamics are more constrained by fuel availability (high $\alpha$) or driven by meteorological forcing (low $\alpha$). Unlike generic self-attention, which treats all spatial positions equally, CAFIM is specifically designed to model the conditional interaction between two physically distinct feature groups.

\paragraph{Decoder and uncertainty estimation.} The decoder consists of three upsampling blocks, each performing $2\times$ bilinear interpolation followed by concatenation with the corresponding CAFIM skip connection and two $3 \times 3$ convolutional layers with batch normalization and ReLU activation. The channel dimensions decrease progressively: $256 \to 128 \to 64 \to 32$. The final feature map passes through Monte Carlo (MC) Dropout ($p=0.3$) before a $1 \times 1$ prediction head. During inference, 20 stochastic forward passes yield a mean prediction and a per-pixel standard deviation, providing calibrated uncertainty estimates that highlight regions where the model is least confident---typically fire perimeter boundaries.

\subsection*{Baseline and comparison architectures}

To contextualize FireSenseNet's performance, we evaluate six additional architectures that systematically vary along two design axes: (1) the degree of inductive bias (CNN $\to$ Transformer) and (2) the presence of explicit cross-modal interaction.

\paragraph{Baseline CNN.} A single-stream CNN encoder-decoder inspired by the architecture of \citet{Huot2022}, where all 12 input channels are concatenated and processed without modality separation. The encoder comprises three stages of dual $3 \times 3$ convolution blocks (channels: $32 \to 64 \to 128$) with max-pooling, a 256-channel bottleneck, and a symmetric decoder with transposed convolutions and skip connections. Our re-implementation adds skip connections and batch normalization relative to the original, which accounts for its higher F1 compared to the published baseline. Parameters: 1.15M.

\paragraph{Pure Transformer (SegFormer).} A SegFormer-style \citep{Xie2021} hierarchical Transformer. Fuel and weather inputs are independently projected via $3 \times 3$ convolutions (each to 32 channels), then concatenated into a 64-channel tensor. The encoder consists of four stages with embedding dimensions $[64, 128, 256, 512]$, each containing two Transformer blocks with Efficient Self-Attention (spatial reduction ratios $[8, 4, 2, 1]$) and Mix-FFN with depthwise $3 \times 3$ convolutions. The decoder follows the SegFormer MLP design: multi-scale features are projected to a common dimension (256) and summed. Parameters: 11.3M.

\paragraph{Small Transformer.} A parameter-reduced SegFormer variant with halved embedding dimensions ($[32, 64, 128, 256]$), reduced block counts ($[1, 1, 2, 1]$ vs.\ $[2, 2, 2, 2]$), and a 128-channel decoder, designed to test whether the full SegFormer's underperformance stems from overfitting due to excess capacity. Parameters: 1.86M.

\paragraph{Regularized Transformer.} Identical to the Small Transformer but with increased MC Dropout ($p=0.4$) and Cutout \citep{DeVries2017} data augmentation (2 holes of size 12 for fuel, 1 hole of size 8 for weather), testing whether aggressive regularization can compensate for limited training data. Parameters: 1.86M.

\paragraph{Hybrid CNN-Transformer.} A two-phase architecture where the first two stages use dual $3 \times 3$ CNN blocks with batch normalization (channels: $32 \to 48 \to 96$) to exploit local inductive biases, and the latter two stages use Transformer blocks (channels: $192, 384$; 2 blocks each with 4 and 8 attention heads) to capture global context. All 12 channels are concatenated at input. Parameters: 5.47M.

\paragraph{Hybrid CNN-CAFIM-Transformer.} Combines dual-branch CNN stems (fuel: $3 \times 3$ convolutions; weather: $5 \times 5$ then $3 \times 3$) with CAFIM fusion at two encoder scales ($64^2$ and $32^2$), followed by two Transformer stages (channels: 192 and 384) for global reasoning. This architecture tests whether adding global attention on top of CAFIM provides incremental benefit. Parameters: 5.94M.

\subsection*{Training strategy}

\paragraph{Addressing extreme class imbalance.} Next-day wildfire prediction exhibits extreme spatial sparsity: typically $<$1\% of pixels show active fire spread. We adopt three complementary strategies. First, we apply Gaussian mixture smoothing ($\sigma \in \{0.4, 0.8\}$) to the PrevFireMask and wind speed channels to diffuse highly localized signals, following \citet{Luo2026}. Second, we employ soft label transformation during training, mapping background pixels to $\mathcal{U}(0.01, 0.03)$ and fire pixels to $\mathcal{U}(0.80, 0.99)$, which prevents over-confidence on hard boundaries and stabilizes convergence. During evaluation, standard hard binary labels (threshold at 0.5) are restored. Third, we use a composite loss function:

\begin{equation}
\mathcal{L} = 0.4 \cdot \mathcal{L}_{\text{BCE}}^{w} + 0.3 \cdot \mathcal{L}_{\text{Dice}} + 0.3 \cdot \mathcal{L}_{\text{Focal}}
\end{equation}

\noindent where $\mathcal{L}_{\text{BCE}}^{w}$ is a weighted binary cross-entropy with positive class weight 3.0, $\mathcal{L}_{\text{Dice}}$ penalizes low spatial overlap, and $\mathcal{L}_{\text{Focal}}$ ($\gamma=2.0$) focuses learning on hard-to-classify boundary pixels. All loss components exclude unknown pixels ($-1$) via a valid mask.

\paragraph{Optimization.} CNN-based models (FireSenseNet and Baseline CNN) use Adam with learning rate $3 \times 10^{-4}$ and batch size 128. Transformer and hybrid variants use AdamW \citep{Loshchilov2019} with learning rate $1 \times 10^{-4}$ (except the Hybrid CNN-CAFIM-Transformer, which uses $3 \times 10^{-4}$), weight decay 0.01, gradient clipping at $\|\nabla\|_{\max}=1.0$, automatic mixed-precision training, and batch size 64. All models employ cosine annealing ($\eta_{\min}=10^{-6}$) over 100 epochs with early stopping (patience 20 for Transformer/hybrid models; 15 for the Baseline CNN). Experiments are conducted on a single NVIDIA GPU; data loading uses 8 parallel workers.

\paragraph{Computational cost.} Table~\ref{tab:complexity} summarizes the computational profile of each architecture. FireSenseNet achieves the best F1 while requiring only moderate FLOPs (2.52G), offering a favorable accuracy--efficiency trade-off. Notably, the SegFormer has comparable FLOPs (2.44G) but nearly 4$\times$ the parameters and 2.2$\times$ the inference latency, reflecting the overhead of self-attention at multiple scales.

\begin{table}[ht]
\centering
\caption{Computational complexity comparison. GFLOPs measured for a single $64 \times 64$ input; inference time averaged over 100 forward passes on a single GPU.}
\label{tab:complexity}
\begin{tabular}{lrrc}
\toprule
\textbf{Model} & \textbf{Params} & \textbf{GFLOPs} & \textbf{Latency (ms)} \\
\midrule
FireSenseNet (CAFIM) & 3.01M & 2.52 & 3.1 \\
Baseline CNN & 1.15M & 0.87 & 1.3 \\
SegFormer & 11.31M & 2.44 & 6.8 \\
Small Transformer & 1.86M & 0.60 & 4.6 \\
Regularized Transformer & 1.86M & 0.60 & 4.6 \\
Hybrid CNN+Trans & 5.47M & 2.20 & 4.1 \\
Hybrid CNN+CAFIM+Trans & 5.94M & 3.46 & 5.0 \\
\bottomrule
\end{tabular}
\end{table}

%% ============================================================
\section*{Results}
%% ============================================================

\subsection*{Architecture comparison}

\begin{figure}[!htbp]
\centering
\includegraphics[width=\linewidth]{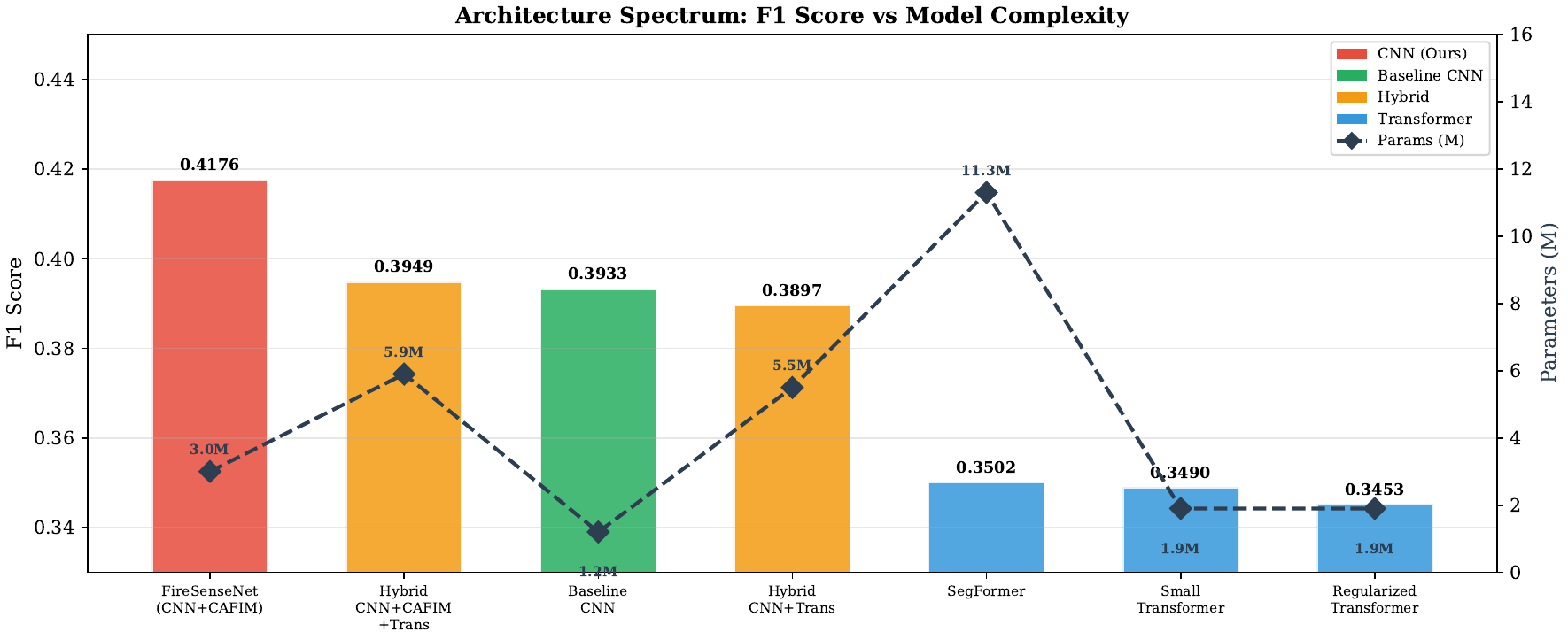}
\caption{Architecture spectrum: F1 score vs.\ model complexity. FireSenseNet achieves the highest F1 (0.4176) with only 3.0M parameters, while Transformer-based architectures cluster at the bottom despite higher parameter counts. The dual y-axis reveals an inverse relationship between F1 and reliance on Transformer components.}
\label{fig:spectrum}
\end{figure}

Table~\ref{tab:main_results} and Figure~\ref{fig:spectrum} present the main results across all seven architectures. A clear performance hierarchy emerges along the CNN--Transformer spectrum.

\begin{table}[ht]
\centering
\caption{Performance comparison on the Next-Day Wildfire Spread test set. All metrics computed under strict next-day-only prediction with $-1$ pixels excluded. Best results in bold.}
\label{tab:main_results}
\begin{tabular}{llrcccc}
\toprule
\textbf{Model} & \textbf{Type} & \textbf{Params} & \textbf{Prec.} & \textbf{Rec.} & \textbf{F1} & \textbf{AUC-PR} \\
\midrule
FireSenseNet (CAFIM) & CNN & 3.0M & \textbf{0.3818} & 0.4609 & \textbf{0.4176} & \textbf{0.3435} \\
Hybrid CNN+CAFIM+Trans & Hybrid & 5.9M & 0.3550 & 0.4449 & 0.3949 & 0.3155 \\
Baseline CNN & CNN & 1.2M & 0.3307 & \textbf{0.4850} & 0.3933 & 0.3354 \\
Hybrid CNN+Trans & Hybrid & 5.5M & 0.3365 & 0.4629 & 0.3897 & 0.3172 \\
SegFormer & Trans & 11.3M & 0.2960 & 0.4287 & 0.3502 & 0.2699 \\
Small Transformer & Trans & 1.9M & 0.2902 & 0.4378 & 0.3490 & 0.2677 \\
Regularized Transformer & Trans & 1.9M & 0.2926 & 0.4211 & 0.3453 & 0.2539 \\
\bottomrule
\end{tabular}
\end{table}

\paragraph{Finding 1: Explicit cross-modal interaction outperforms all alternatives.} FireSenseNet achieves the highest F1 (0.4176) and AUC-PR (0.3435) among all tested architectures, surpassing the published baseline of \citet{Huot2022} (F1 = 0.359) by 16.3\%. Even compared to our stronger re-implemented Baseline CNN (F1 = 0.3933), the 6.2\% relative improvement demonstrates that separating fuel and weather processing and fusing them through learned attention provides genuine architectural benefit beyond increased model capacity.

\paragraph{Finding 2: Transformer components consistently degrade performance.} A striking monotonic trend emerges: as architectures incorporate more Transformer components, performance decreases. The Hybrid CNN+CAFIM+Transformer (F1 = 0.3949) falls below pure-CNN FireSenseNet despite nearly doubling the parameter count. The full SegFormer performs worst among all non-regularized models (F1 = 0.3502) despite having the most parameters (11.3M)---a 3.8$\times$ parameter disadvantage relative to FireSenseNet.

\paragraph{Finding 3: Transformer underperformance is structural, not due to overfitting.} If the Transformer's poor performance were caused by overfitting from excess capacity, reducing parameters should help. The Small Transformer (1.9M parameters) performs almost identically to the full SegFormer (11.3M), ruling out this explanation. Furthermore, adding aggressive regularization (Dropout 0.4, Cutout augmentation) in the Regularized Transformer marginally \emph{worsens} performance---likely because randomly masking input regions destroys the sparse but informationally critical fire mask signals. The Transformer's underperformance is thus attributable to the lack of suitable inductive biases (translation equivariance, locality) for this small-data, spatially sparse task.

\begin{figure}[!htbp]
\centering
\includegraphics[width=\linewidth]{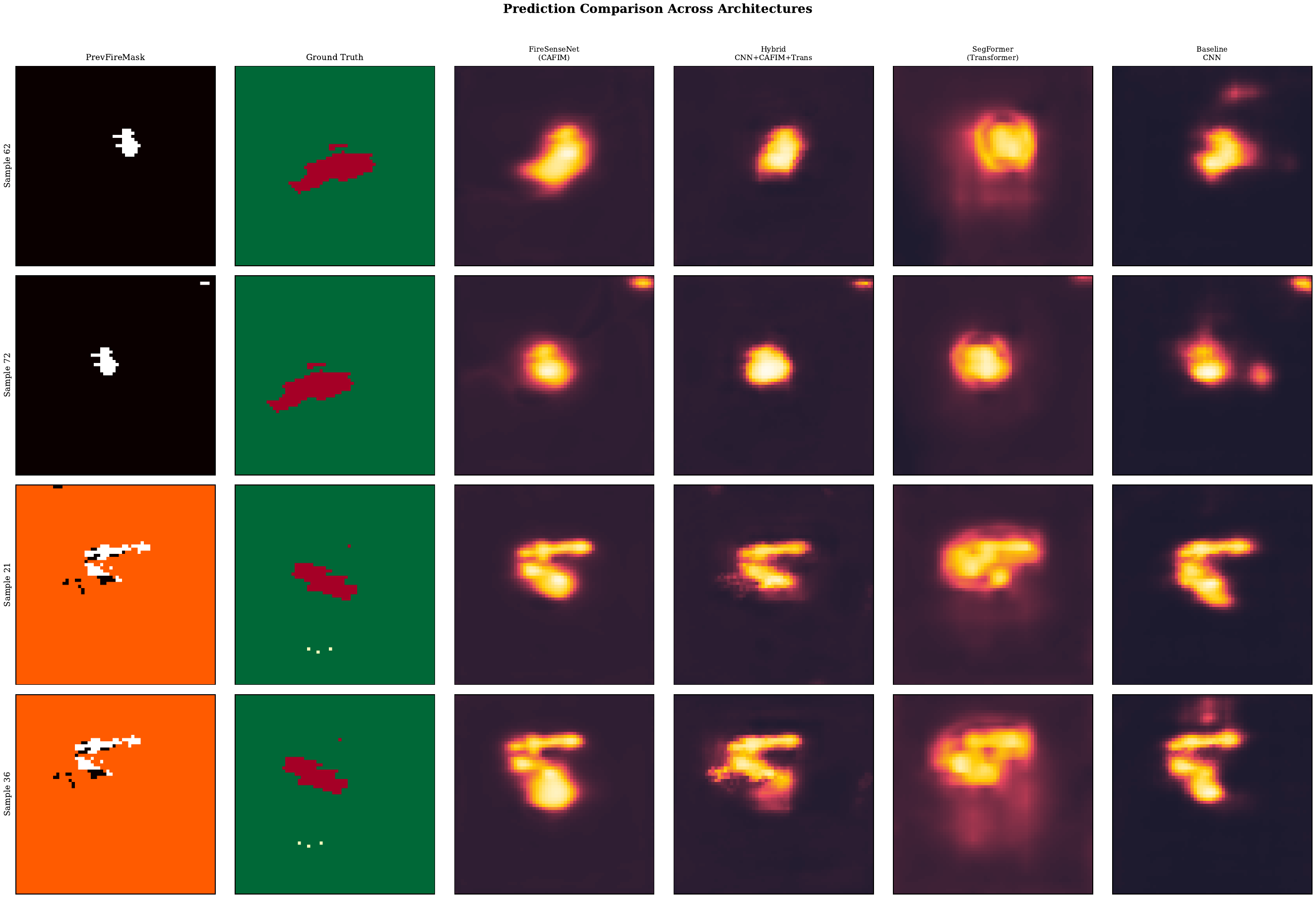}
\caption{Qualitative prediction comparison on four test samples. FireSenseNet (CAFIM) produces spatially precise predictions matching ground truth geometries, while the SegFormer generates diffuse, over-spread probability maps. The Baseline CNN captures fire locations but with less boundary precision.}
\label{fig:prediction_comparison}
\end{figure}

\paragraph{Qualitative comparison.} Figure~\ref{fig:prediction_comparison} shows predictions from four representative architectures on test samples with active fire spread. FireSenseNet produces the most spatially concentrated predictions that closely match the ground truth fire geometries. The SegFormer generates diffuse, over-spread predictions---a consequence of its global attention mechanism distributing probability mass across the entire spatial extent. The Baseline CNN captures fire locations but with lower precision at boundaries.

\subsection*{Ablation and diagnostic analyses}

We investigate \emph{why} FireSenseNet outperforms alternatives through two complementary lenses: a module-level ablation that isolates the contribution of CAFIM, and a channel-level analysis that reveals which input features the model relies on.

\begin{figure}[!htbp]
\centering
\includegraphics[width=\linewidth]{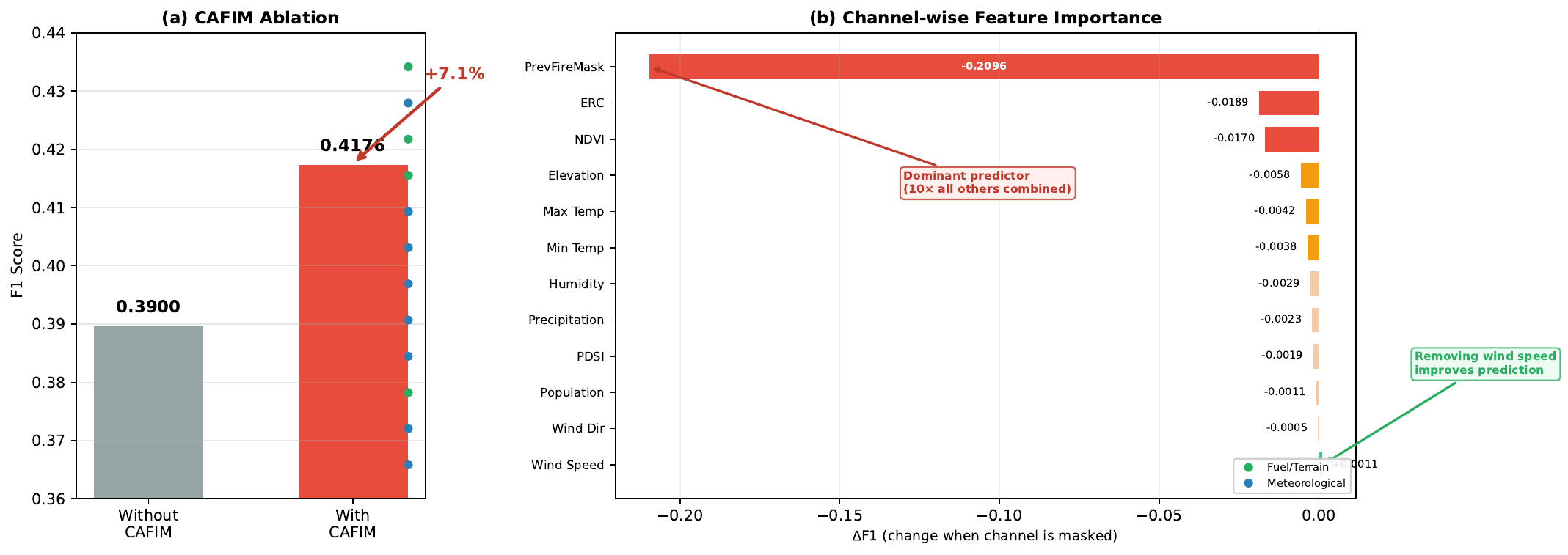}
\caption{(a) CAFIM ablation: removing CAFIM and replacing it with simple concatenation reduces F1 by 7.1\%. (b) Channel-wise feature importance via systematic ablation. PrevFireMask dominates ($\Delta$F1 = $-0.21$); wind speed is the only channel whose removal \emph{improves} performance (green bar), indicating it acts as noise at this temporal resolution.}
\label{fig:ablation_and_features}
\end{figure}

\begin{figure}[!htbp]
\centering
\includegraphics[width=\linewidth]{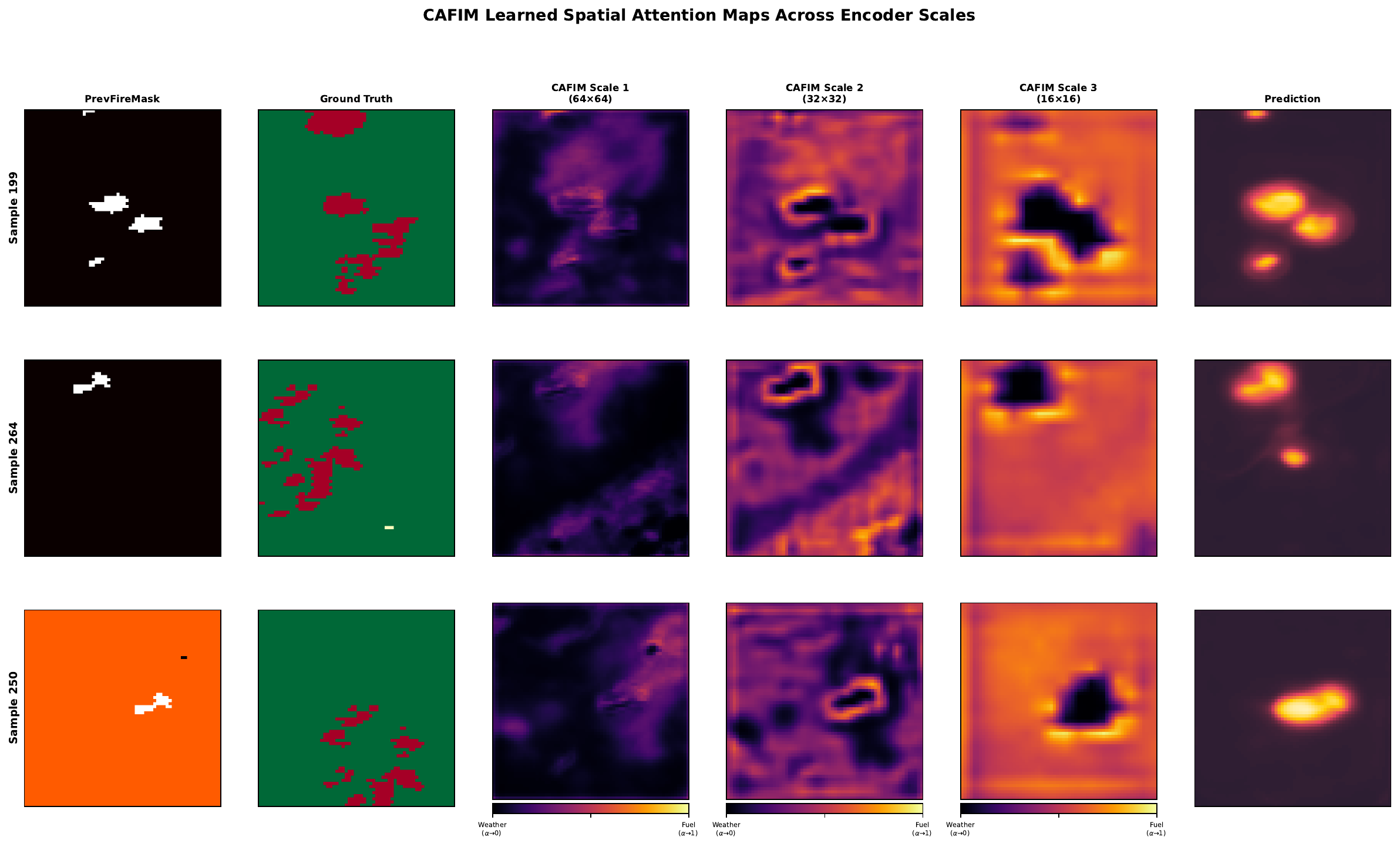}
\caption{CAFIM attention maps at three encoder scales for representative test samples. The learned spatial attention $\alpha$ determines the relative contribution of fuel features ($\alpha \to 1$, yellow) vs.\ weather features ($\alpha \to 0$, dark). At fine scales, CAFIM focuses sharply on fire perimeters; at coarse scales, it captures broader susceptibility patterns.}
\label{fig:cafim_attention}
\end{figure}

\paragraph{CAFIM ablation.} To separate the effect of cross-modal attention from the dual-branch design itself, we trained an ablation variant that replaces CAFIM with simple channel-wise concatenation while keeping both encoder branches intact. This ablated model achieves F1 $\approx$ 0.39, compared to 0.4176 for the full FireSenseNet---a relative improvement of 7.1\% attributable to CAFIM (Figure~\ref{fig:ablation_and_features}a). This confirms that explicitly modeling the spatial interaction between fuel and weather features is more effective than letting the decoder implicitly learn these relationships through concatenated skip connections. Figure~\ref{fig:cafim_attention} provides qualitative evidence: the learned attention maps exhibit sharp spatial focus on fire perimeters at fine scales ($64 \times 64$), broadening to regional susceptibility patterns at coarser scales---consistently highlighting the transition zone where fuel-weather interplay determines whether fire advances.

\paragraph{Channel-wise feature importance.} We complement the module-level analysis with a systematic channel masking experiment: each input channel is replaced by its global mean and the resulting $\Delta$F1 is measured (Table~\ref{tab:features}, Figure~\ref{fig:ablation_and_features}b). The results reveal a steep importance hierarchy. \textbf{PrevFireMask} is overwhelmingly dominant ($\Delta$F1 = $-0.21$), confirming that next-day spread prediction is fundamentally a short-term persistence problem. Among secondary features, \textbf{ERC}---an integrated measure of fire potential---is the most influential weather variable ($\Delta$F1 = $-0.019$), followed by \textbf{NDVI} ($\Delta$F1 = $-0.017$), consistent with vegetation serving as the primary fuel source. The most striking finding is that \textbf{wind speed exhibits a positive $\Delta$F1 of $+0.001$}: masking it actually \emph{improves} prediction. This counter-intuitive result reflects a mismatch between daily-averaged wind data and the instantaneous gusts that drive real fire spread, suggesting that incorporating temporally resolved wind observations could substantially improve future models.

\begin{table}[ht]
\centering
\caption{Feature importance via channel masking on FireSenseNet. $\Delta$F1 indicates the change when the channel is replaced by its global mean.}
\label{tab:features}
\begin{tabular}{llrl}
\toprule
\textbf{Channel} & \textbf{Group} & \textbf{$\Delta$F1} & \textbf{Interpretation} \\
\midrule
PrevFireMask & Fuel & $-0.2096$ & Dominant predictor; fire persistence \\
ERC & Weather & $-0.0189$ & Integrated fire potential index \\
NDVI & Fuel & $-0.0170$ & Vegetation fuel availability \\
Elevation & Fuel & $-0.0058$ & Topographic channeling \\
Wind Speed & Weather & $+0.0011$ & Acts as noise (coarse resolution) \\
\bottomrule
\end{tabular}
\end{table}

\subsection*{Evaluation integrity and uncertainty}

Beyond prediction accuracy, we examine two dimensions of result quality: whether reported scores faithfully reflect model capability, and whether the model can communicate its own confidence.

\begin{figure}[!htbp]
\centering
\includegraphics[width=\linewidth]{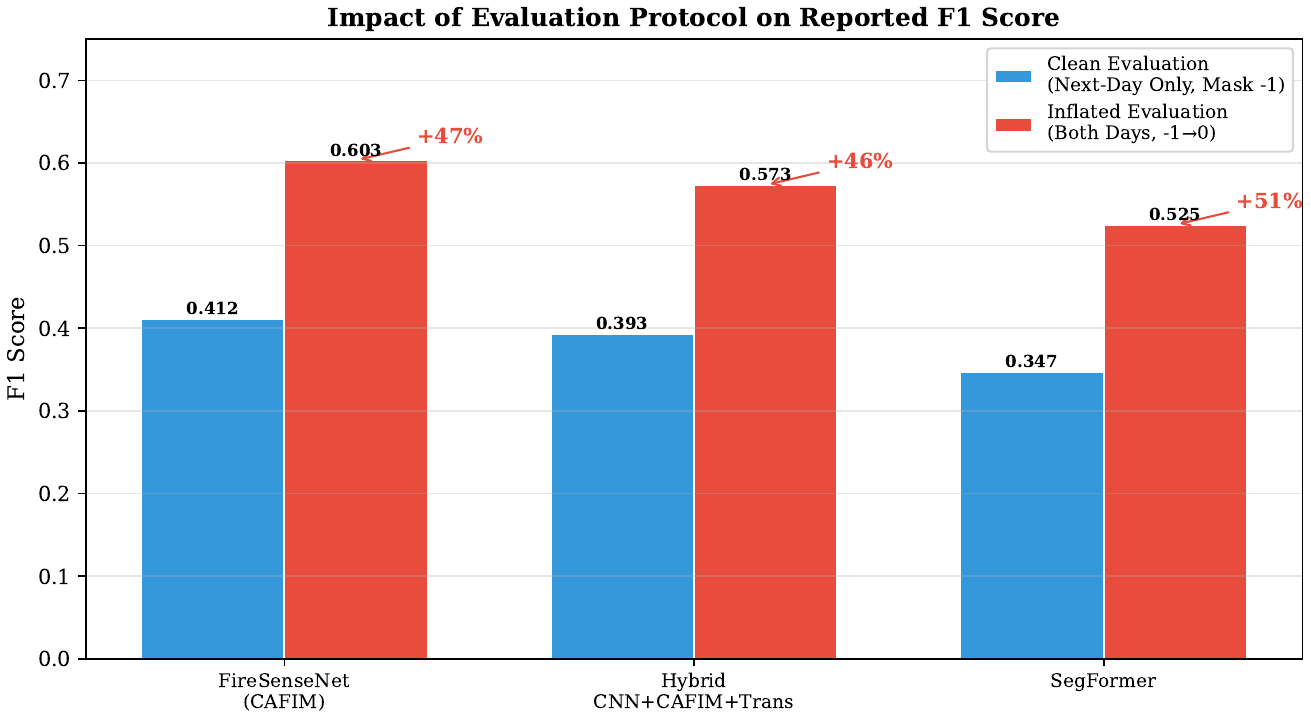}
\caption{Clean vs.\ inflated evaluation protocols. Including previous-day fire in the target and relabeling unknown pixels as non-fire inflates reported F1 by 44--50\%, disproportionately benefiting weaker architectures and masking genuine performance differences.}
\label{fig:eval_inflation}
\end{figure}

\paragraph{Evaluation protocol sensitivity.} To quantify the impact of evaluation methodology, we re-evaluated three architectures under an alternative protocol that (i) includes the previous-day fire mask in the target (``Both Days'' prediction) and (ii) treats unknown pixels ($-1$) as non-fire rather than excluding them. The inflated protocol boosts reported F1 by 44--50\% across all architectures (Table~\ref{tab:eval_protocol}, Figure~\ref{fig:eval_inflation}). The inflation is mechanical: including PrevFireMask in the target allows models to achieve high true positive rates by simply reproducing their input, while relabeling unknown pixels inflates the true negative count. Critically, the inflation disproportionately benefits weaker models---the SegFormer's gap to FireSenseNet shrinks from 19.2\% under clean evaluation to 14.9\% under inflated evaluation---masking genuine architectural differences. This analysis highlights the sensitivity of reported metrics to evaluation choices and underscores the need for standardized protocols. We note that these inflation effects are methodological rather than model-specific, and affect all architectures similarly.

\begin{table}[ht]
\centering
\caption{Impact of evaluation protocol on reported F1. ``Clean'' uses next-day-only prediction with $-1$ exclusion; ``Inflated'' includes previous-day fire in the target and sets $-1 \to 0$.}
\label{tab:eval_protocol}
\begin{tabular}{lccc}
\toprule
\textbf{Model} & \textbf{Clean F1} & \textbf{Inflated F1} & \textbf{Inflation} \\
\midrule
FireSenseNet (CAFIM) & 0.4176 & 0.6033 & +44.4\% \\
Hybrid CNN+CAFIM+Trans & 0.3949 & 0.5733 & +45.2\% \\
SegFormer & 0.3502 & 0.5252 & +50.0\% \\
\bottomrule
\end{tabular}
\end{table}

\begin{figure*}[!htbp]
\centering
\includegraphics[width=\textwidth]{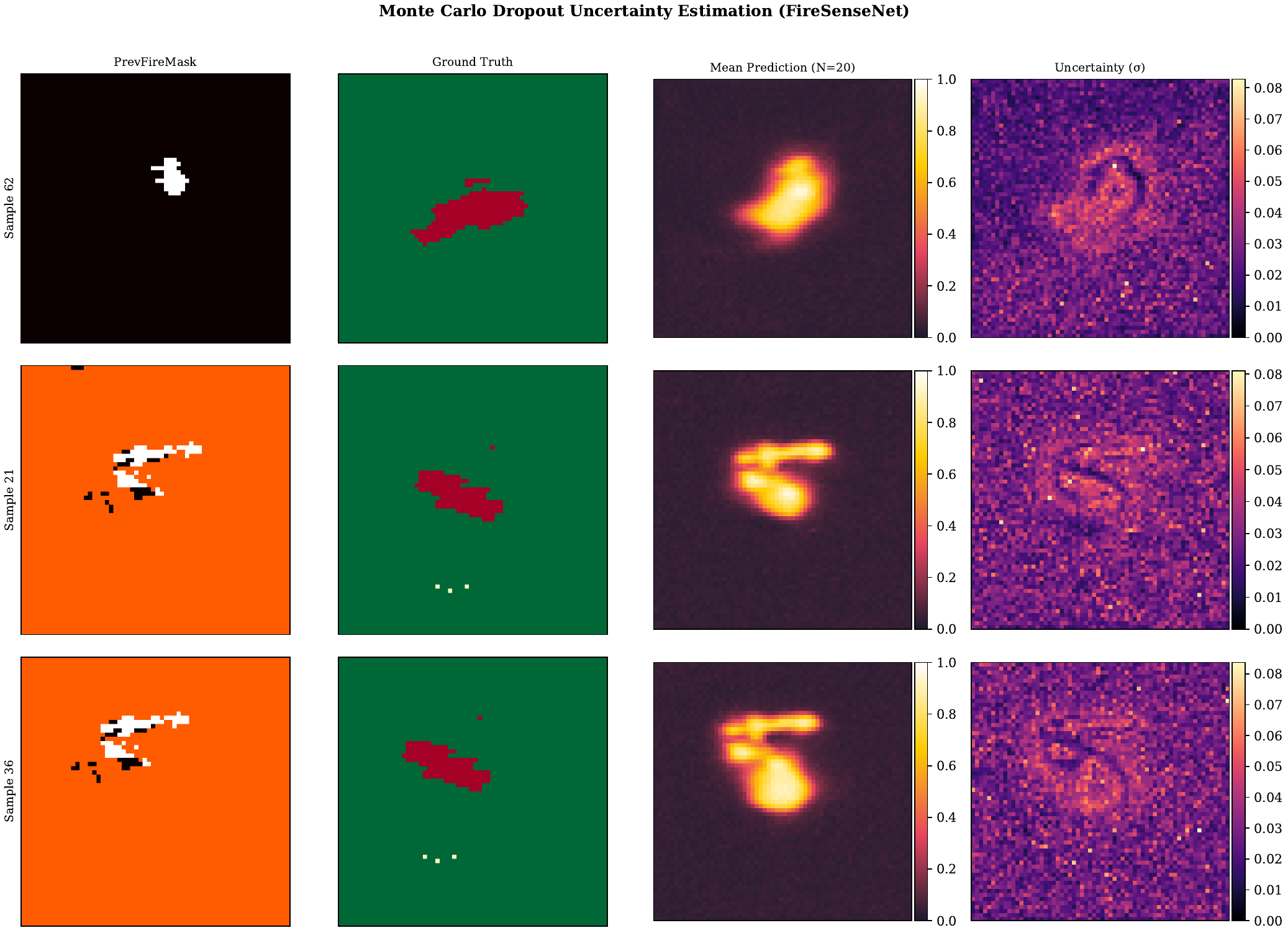}
\caption{Monte Carlo Dropout uncertainty estimation for three test samples. From left to right: previous-day fire mask, ground truth, mean prediction (20 stochastic passes), and pixel-level uncertainty ($\sigma$). Uncertainty concentrates at fire perimeter boundaries, providing actionable information for prioritizing ground-truth verification during emergency response.}
\label{fig:uncertainty}
\end{figure*}

\paragraph{Pixel-level uncertainty estimation.} Using Monte Carlo Dropout with 20 forward passes, FireSenseNet produces calibrated uncertainty maps alongside predictions (Figure~\ref{fig:uncertainty}). High-uncertainty regions consistently correspond to fire perimeter boundaries---the dynamic frontier where the model is least confident about whether fire will advance. This spatial correspondence is physically meaningful: perimeter pixels represent the transition between ``certainly burning'' and ``certainly safe,'' making the uncertainty maps directly actionable for emergency responders who must prioritize ground-truth verification in resource-constrained settings.

%% ============================================================
\section*{Discussion}
%% ============================================================

\subsection*{Why CNNs outperform Transformers for wildfire spread prediction}

Our results reveal a clear performance gradient that \emph{decreases} with increasing reliance on Transformer components---a finding that challenges the prevailing trend toward attention-based architectures in remote sensing. We attribute this to three factors operating in concert.

\textbf{Data scarcity favors inductive bias.} The dataset contains approximately 15,000 training samples of $64 \times 64$ pixels. At this scale, CNN inductive biases---translation equivariance and locality---provide strong, beneficial priors that compensate for limited training data. Transformers, lacking these priors, must learn basic spatial regularities from scratch, expending capacity on patterns that CNNs obtain for free. Our results confirm the finding of \citet{Dosovitskiy2021}: Transformer advantages emerge at large dataset scales, not small ones.

\textbf{Informative signals are local, not global.} The channel-wise feature importance analysis reveals that fire spread is dominated by the previous-day fire mask---a signal with sharp, localized spatial structure concentrated at existing fire perimeters. The secondary predictors (ERC, NDVI, Elevation) also operate at local to medium-range spatial scales well within the effective receptive field of multi-scale CNNs. The global attention mechanism of Transformers, while theoretically capable of modeling long-range dependencies, provides diminishing returns when the most informative signals are spatially concentrated.

\textbf{Cross-modal structure is better modeled by dedicated mechanisms.} The 7.1\% relative F1 gain from CAFIM over simple concatenation demonstrates that the interaction between fuel and weather modalities has exploitable structure that benefits from dedicated architectural support. Generic self-attention treats all positions and channels uniformly; CAFIM, by contrast, enforces a physically motivated complementary gating that explicitly partitions the feature space along modal boundaries. This targeted inductive bias is more efficient than learning the same structure through generic attention.

These three factors are mutually reinforcing but also context-dependent. On larger-scale datasets, Transformers have matched or outperformed CNNs across vision tasks \citep{Dosovitskiy2021}; our finding does not contradict this broader trend but rather delineates the regime boundary. When training data is limited, labels are sparse, and domain-specific structure is available, encoding that structure into the architecture yields greater returns than relying on the flexibility of general-purpose attention.

%% ============================================================
\section*{Conclusions}
%% ============================================================

We presented FireSenseNet, a dual-branch CNN with Cross-Attentive Feature Interaction Modules for next-day wildfire spread prediction, alongside a systematic architecture comparison spanning the CNN--Transformer spectrum. Three principal findings emerge. First, on the Google Next-Day Wildfire Spread benchmark, explicit cross-modal attention (CAFIM) between fuel/terrain and meteorological features yields the highest performance (F1 = 0.4176), outperforming both single-stream CNNs and Transformer-based architectures with up to 3.8$\times$ more parameters. Second, channel-wise feature importance analysis reveals that prediction is dominated by short-term fire persistence (PrevFireMask), while wind speed---conventionally considered a key fire weather variable---acts as noise at the daily temporal resolution of current datasets. Third, our evaluation protocol analysis demonstrates that common evaluation choices can inflate reported F1 scores by over 44\%, underscoring the need for standardized benchmarking.

These results suggest that for small-scale, spatially sparse remote sensing prediction tasks, encoding domain knowledge in architecture design---through modality-aware processing and explicit cross-modal interaction---is more effective than relying on the representational flexibility of Transformers or the brute-force capacity of larger models. This study is limited to a single benchmark dataset with fixed spatial resolution ($64 \times 64$) and daily temporal granularity. Future work should validate these findings on multi-temporal datasets \citep{Gerard2023} where Transformer architectures may prove more competitive, incorporate temporally resolved wind observations to unlock currently noisy channels, and extend CAFIM to model temporal cross-modal interactions in multi-day prediction scenarios.

\section*{Acknowledgments}

This work was supported by the Korea Institute of Energy Technology Evaluation and Planning (KETEP) and the Ministry of Climate, Energy, Environment (MCEE) of the Republic of Korea (No.\ RS-2025-07852969). This research was also supported by the 2023-MOIS36-004 (RS-2023-00248092) of the Technology Development Program on Disaster Restoration Capacity Building and Strengthening funded by the Ministry of Interior and Safety (MOIS, Korea).

\section*{Data Availability}

The Next-Day Wildfire Spread dataset is publicly available through Google Research \citep{Huot2022}. Code and trained model weights will be released upon publication.

\bibliography{references}

\end{document}